\documentclass[fleqn,10pt]{olplainarticle}
\usepackage{orcidlink}
\usepackage{multirow}
\usepackage{subfigure}
\hypersetup{
    hidelinks
}

\title{Clustering algorithms for multivariate wind farm SCADA data filtering}

\author{Nicolò Italiano\orcidlink{0000-0002-9072-8534}}
\author{Vasilis Pettas\orcidlink{0000-0001-9985-9031}}
\author{Tuhfe Göçmen\orcidlink{0000-0003-2825-518X}}
\author{Nicolaos A. Cutululis\orcidlink{0000-0003-2438-1429}} 

\affil{Department of Wind and Energy Systems, Technical University of Denmark, Frederiksborgvej 399, Roskilde, 4000, Denmark}


\begin{abstract}
During wind farm operation, Supervisory Control and Data Acquisition (SCADA) systems record numerous anomalies, transients, and specific operational modes, leading to large datasets. However, for a wide range of applications, only measurements corresponding to normal operation are required and, therefore, the SCADA data must be filtered. For this purpose, several methods have been proposed to automate and replace manual filtering conducted by experts via visual inspection of the data. In this paper, we compare the filtering accuracy of multiple clustering algorithms against manual filtering, introducing evaluation metrics that are suitable for unlabeled data and robust across potential applications. Based on the results, we provide recommendations for generalizing model calibration to different datasets and discuss potential use cases for each model.  The models are applied to the SCADA data of three turbines of an existing offshore wind farm, using 10-minute statistics across multiple data channels. In addition to the anomalies and operational modes typically recorded, the dataset presents a large number of non-evident outliers due to several field tests. Overall, the results highlight the importance of extending the analysis beyond the power curve, both in feature selection and in the design of evaluation metrics. In most cases, cluster-based methods are able to detect both evident and subtle outliers, achieving higher accuracy than manual filtering. However, the accuracy and the amount of data retained vary considerably depending on the model, and expert involvement remains necessary, though to a reduced extent compared to manual filtering. 
\end{abstract}

\begin{document}

\flushbottom
\maketitle
\thispagestyle{empty}

\section{Introduction}
\label{sec1}

During their operational lifetime, wind farms are exposed to a variety of non-stationary atmospheric conditions, which may vary considerably among wind turbines (WT), due, for example, to the wake effects or to the heterogeneous spatial distribution of wind resources at the site. Moreover, WTs are subject to different operation modes and transients, e.g., idling, start-up / shut-down, and curtailment. All this has an impact not only on the power performance of each single WT, but also on their respective structural loads and operational lifetime. Therefore, it is paramount to continuously monitor, analyze, and control the operation of each WT. This is done using a Supervisory Control And Data Acquisition (SCADA) system, collecting a wide range of data channels including weather-related variables, such as local wind speed and direction, outdoor temperature, and operational parameters, e.g., power generated, blade pitch angle, rotor speed, and additional data according to the sensors installed. The SCADA system typically stores and transmits 10-minute averaged data, although the actual measurements are collected at a higher sampling frequency \cite{pandit_scada_2023}, resulting in a significant volume of data. The data can be used for a wide range of applications, including, among others, fault detection, condition-based monitoring for predictive maintenance, northing calibration, power curve modeling, normal behavior modeling, power forecasting, and wake model parameter calibration. 
 
However, many of these applications require input data corresponding exclusively to the Maximum Power Point Tracking (MPPT), hereafter referred to as the normal operation, for clarity. For instance, modeling the power curve based on SCADA data containing idling or curtailment periods would result in an underestimation of the power performance. The same applies to training a model for power forecasting. Yet, in practice, SCADA systems record a large number of outliers, anomalies, transients, and specific operational modes deviating from the normal operation, typically without any corresponding labels or flags. The traditional approach to filtering this data relies on an in-depth data analysis conducted by dedicated experts, which defines a set of WT-specific rule-based filters, e.g., removing negative power values and pitch angles deviating from the fine pitch at below-rated wind speeds. However, such a process is time-consuming and resource-intensive, in particular for large-scale wind farms. 

In recent years, several studies have explored different approaches to facilitate, improve, and increase the level of automation in SCADA data filtering of wind farms. The methods can be divided into three main categories \cite{wang, kijanowski_cluster-based_2025}: i) statistical, ii) image-based, and iii) machine-learning (ML) approaches, with the first and the last often coupled. The first methods consist of applying filters based on the statistical distributions of one or more data signals. For example, the inter-quartile range (IQR) is commonly used to filter out data points outside the first and third quartiles of the power curve \cite{zhao, luo_method_2022, wang_pre-filtering_2025}, while the Mahalanobis distance works similarly, but takes into account the covariance between variables \cite{kijanowski_cluster-based_2025, vasquez-rodriguez_anomaly-based_2024}. A major limitation with these approaches is that the filtering thresholds must be adapted to the data spread, which depends on the turbulence characteristics of the site considered \cite{alcayaga_filtering_2020}, and that high concentrations of outliers can significantly skew the statistical distributions. The category ii), i.e., image processing techniques, have also been investigated. Long et al. \cite{long_image-based_2020} applied mathematical morphology operations to a binary image of the power curve, achieving a higher abnormal data deletion rate than other statistical and ML methods. Similarly, Wang et al. \cite{wang_fast_2021} filtered the data based on the pixel spatial distribution of the power curve image, outperforming the algorithm of \cite{long_image-based_2020} in both accuracy and computational time. Yet, image-based methods have been mainly applied to the power curve, neglecting the multivariate dependence of the power on both ambient conditions and operational parameters \cite{astolfi}. The third category, ML algorithms, can be applied to capture the underlying correlations between SCADA data signals. Most studies have focused on clustering algorithms, suitable for the unlabeled data typically generated by SCADA systems. 
Zhao et al. \cite{zhao} and Wang et al. \cite{wang} proposed methods based on Density-Based Spatial Clustering of Applications with Noise (DBSCAN) to eliminate stacked outliers from the power curve, corresponding to curtailments, after applying IQR filtering to filter out sparse outliers. However, the studies used exclusively the 10-minute-averaged power and local wind speed as input. Kijanowski et al. \cite{kijanowski_cluster-based_2025} combined K-Means++ clustering, used to partition the power curve, with the Mahalanobis distance for outlier removal, achieving similar performance of DBSCAN. Morrison et al. \cite{MORRISON2022473} introduced additional inputs, including the mean values of blade pitch angle and temperature, and compared the accuracy of several statistical-based and ML models in detecting sparse outliers, curtailments, and idling periods. The Gaussian Mixture Model (GMM) proved to be the most effective method, in particular after manually filtering out the most evident anomalies. However, the outputs of the different models were not compared with a thorough manual filtering.

Although several algorithms and methods have been proposed and compared in the literature, there is no consensus on which approach is most suitable for filtering SCADA data. 
This is partly due to the heterogeneity of evaluation metrics used, which hinders meaningful and reliable comparisons. Several studies proposed generalized metrics, such as the silhouette score and Spearman's correlation \cite{wang, exploratory}. Others compared the accuracy of the power prediction of regression models trained on both raw and filtered data \cite{zhang_wind_2024, lin_wind_2020}. While these metrics can be applied to unlabeled data, they are either too generic, evaluating only the quality of the clusters, or tailored to specific applications, e.g., power forecasting. This limits their relevance for other applications, such as normal behavior modeling and structural loads forecasting. A few studies introduced metrics targeted to the power curve, such as the power curve modeling error in \cite{zhao}, and the change to wind speed inter-quartile ranges in \cite{MORRISON2022473}, together with the elimination rate, which gives an indication of how much data is filtered out by a model. This work builds upon these evaluation metrics, broadening the scope to other operational curves to better address the multivariate nature of SCADA data and its wide range of applications.

Besides this, the literature often focuses on comparing filtering methods, without evaluating their accuracy relative to manual filtering methods, nor discussing the challenges and implications of replacing them. Moreover, the SCADA data of the studies analyzed presented mostly visible anomalies and outliers that are detectable via visual inspection of the power curve, such as idling periods, curtailments, and sparse outliers. This simplifies the task for the filtering methods and may not reflect their accuracy on complex datasets.

In this study, we compare the output from filtering using multiple clustering algorithms with manual filtering by visual inspection, described in section \ref{subsect2.2}. Specifically, the study will focus on GMM and DBSCAN, which have been proven effective for this purpose, as well as Hierarchical DBSCAN (HDBSCAN). The latter has shown higher accuracy than DBSCAN across different applications, e.g., \cite{he_oct_2024}, but, to the authors' knowledge, it has not been used for SCADA data filtering in wind farms.
The algorithms are applied to the SCADA data of an existing wind farm, which has been subjected to several wind
farm flow control experiments and other field tests \cite{bossanyi_full-scale_2023}, resulting in a significant amount of data, which is often not easily detectable through visual inspection. Finally, motivated by recent work encouraging analysis beyond the commonly used power and wind‑speed mean signals \cite{astolfi, MORRISON2022473}, a broad set of SCADA channels is investigated as input features, including multiple 10‑minute statistics such as maximum, minimum, and standard deviation. 

To summarize, the contributions of the study are the following:
\begin{itemize}
\item[-] Benchmarking multiple cluster-based SCADA data filtering methods against manual filtering by visual inspection, using a case study characterized by subtle and challenging data deviating from normal operation;
\item[-] Defining robust evaluation metrics for unlabeled data, targeted to WT operational curves, reusable for different applications;
\item[-] Apply HDBSCAN to wind farm SCADA data filtering.
\end{itemize}

The data used in the study, as well as the methodology for performing the manual (baseline) and cluster-based filtering, are presented in Section \ref{sec2}. Section \ref{sec3} describes the evaluation metrics used to compare the performance of the different filtering approaches. The results are presented and discussed in Section \ref{sec4}, focusing on model comparison, potential applications, and limitations of the study. Finally, Section \ref{sec5} concludes the paper.

\section{Methodology}
\label{sec2}

\begin{figure}[htbp]
\centering
\includegraphics[width=\textwidth]{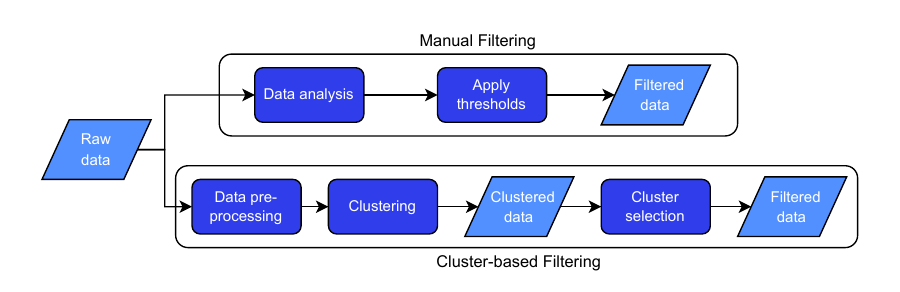}
\caption{Workflow for manual filtering (baseline) and cluster-based filtering. Parallelograms represent input/output data, while rounded rectangles indicate processing steps.}\label{fig:Figure_1}
\end{figure}

An overview of the workflow is illustrated in Figure \ref{fig:Figure_1}, highlighting the differences between the filtering approaches. For the manual filtering, the raw data of each WT is analyzed and visually inspected to define threshold-based filters, as described in detail in Section \ref{subsect2.2}. For the cluster-based filtering, the data is pre-processed, e.g., standardized, before applying the clustering algorithms, which will classify the data into different groups of points, i.e., clusters. Then, a final step is required to select the cluster(s) corresponding to the WT normal operation, as further detailed in Section \ref{subsect2.3}. Finally, the accuracy of the baseline and cluster-based filtering with different algorithms is evaluated and compared using the metrics defined in Section \ref{sec3}.

\subsection{Wind farm data}
\label{subsect2.1}

The available data consists of 10-minute-averaged statistics of SCADA measurements collected at the Lillgrund Wind Farm, comprising 48 SWT-2.3-93 WTs with a total installed capacity of 110.4 MW, located offshore in the Øresund strait, in southern Sweden. The dataset features multiple channels, including active power, wind speed measured with the nacelle anemometer, nacelle orientation, blade pitch angle, and generator speed. However, it does not contain any flags indicating whether the WT is operating in normal power production mode. Northing calibration has been applied to correct the nacelle orientation signal using FLASC \cite{flasc2026}, an open-source tool to support SCADA data filtering and analysis. An overview of the available channels for all WTs is listed in Table \ref{tab:signals}.

\begin{table}[htbp]
\centering
\begin{tabular}{l c c c c}
\hline
  Variable & Mean & Max & Min & Std \\ \hline 
  Active Power & X & X & X & X \\
  Wind Speed & X & X & X & X \\
  Nacelle Orientation & X & N/A & N/A & X \\
  Pitch Angle & X & X & X & X \\
  Generator Speed & X & X & X & X \\
\end{tabular}
\caption{SCADA data channels available in the study. X indicates availability, N/A indicates that the corresponding statistic is not applicable for the given variable.}\label{Table_1}
\label{tab:signals}
\end{table}

The data cover more than 3 years of operations for several wind turbines, recorded between 2019 and 2022, during which the wind farm has undergone multiple field tests. This includes, for example, wake-steering campaigns, in which clockwise and anticlockwise yaw misalignments were applied to some WTs \cite{wake_steering}, and axial-induction control campaigns, characterized by toggling tests between base and down-regulated control for specific WT rows and wind direction sectors \cite{bossanyi_full-scale_2023}. These experiments enriched the dataset with a great number of data points that, although not representative of normal WT operation, are challenging to detect, as further explained in Section \ref{subsect2.2}. 

\begin{figure}[htbp]
    \centering
    \subfigure{
        \includegraphics[width=0.45\textwidth]{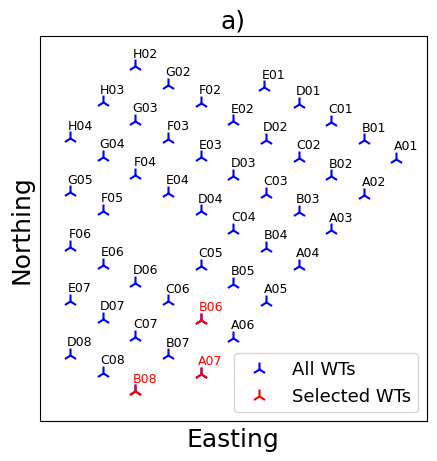}
    }
    \hfill
    \subfigure{
        \includegraphics[width=0.45\textwidth]{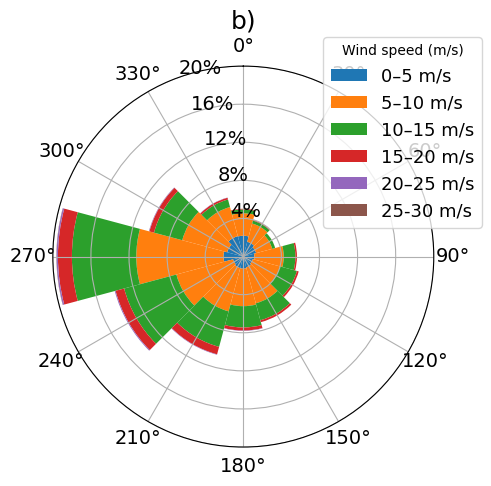}
    }
    \caption{Wind farm layout (a), and SCADA-derived wind rose (b) of Lillgrund wind farm.}
    \label{fig:lillgrund}
\end{figure}

To limit the amount of datasets to be processed and manually filtered, a subset of 3 WTs is selected as the main sample for the data filtering, as highlighted in red in Figure \ref{fig:lillgrund}(a), where the layout of Lillgrund offshore wind farm is presented, alongside the wind rose in Figure \ref{fig:lillgrund}(b), estimated from the SCADA data using measurements from nacelle orientation and nacelle anemometers of the upstream turbines per sector. These specific WTs were selected based on the following criteria: i) data availability, to choose among the units with the largest measurement periods in the dataset, ii) participation in field tests and/or curtailments, and iii) including WTs with different wake conditions, with B06 operating frequently in the wakes of other WTs, while B08 experiences mostly free‑flow conditions. The three datasets present different characteristics. A07 has the fewest valid timestamps (64,424) but shows several curtailments and downregulations for extended periods. The data for B08 and B06 cover a larger period, 150,271 and 142,127 valid timestamps, respectively, with fewer curtailments and other operations deviating from normal behavior.

\subsection{Manual filtering}
\label{subsect2.2}

To perform the manual filtering (baseline), the SCADA data of the 3 WTs is thoroughly analyzed, with particular attention to the main operational curves, i.e., power curve, blade pitch angle curve, and generator speed curve. Then, we systematically define and apply the thresholds listed in Table \ref{tab:Table_2} to label and remove anomalies and operational modes deviating from normal operation. The entire baseline filtering process is carried out using FLASC \cite{flasc2026}, which we extended with additional tailored filters.
Specifically, we filter the data points where the wind speed is close to the cut-off, but the power is below rated, corresponding to shut-downs or down-regulation at high-wind speeds. Then, curtailments are flagged when the mean value of the power is below-rated, and the standard deviation (\textit{std}) is below a specific value, indicating constant power production. Setting the \textit{std} threshold too high would include normal operation cases in which the variations in power were not significant over a 10-minute interval, while a low threshold may not cover the entire periods during which the WT is curtailed. Similarly, to detect operational modes in which the nominal control setpoints are altered, e.g., down-regulation due to axial-induction control, we select the data below rated power with a constant mean pitch angle different from the fine-pitch values corresponding to normal operation. Also in this case, we define a \textit{std} threshold for the pitch, to isolate the periods in which it remained constant. Afterward, to identify idling and start-up/shut-down transients, we capped the mean and then the minimum power at a value close to zero, which was adjusted to be suitable for all WTs, as they have different measurement noise and sensor precision. Pitch angles above 30° were also removed as an additional measure against idling and other atypical operating conditions. Finally, to eliminate sparse outliers, we apply IQR filtering to the power curve, setting percentile thresholds based on each WT's data spread, which varies considerably across datasets. 
As several filter thresholds were WT-specific and depended, among other things, on the fine-pitch settings used, as well as the noise and standard deviation of the power and pitch signals, they were defined by visual inspection of each dataset, following a cumbersome and iterative process. 

\begin{table}[ht]
\centering
\resizebox{\textwidth}{!}{%
\begin{tabular}{l cc cc ccc}
\hline
 & \multicolumn{2}{c}{Wind Speed (m/s)} 
 & \multicolumn{3}{c}{Power (kW)} 
 & \multicolumn{2}{c}{Pitch Angle (°)} 
 \\
Label & Mean & Std & Mean & Std & Min & Mean & Std \\ \hline

Cut-off 
  & $>20$ & -- 
  & $<2275$& -- & --
  & -- & -- 
  \\

Curtailment 
  & -- & -- 
  & $<2275$& $<25^*$ & --
  & -- & --
  \\

Down-regulation  
  & -- & -- 
  & $< 2275$ & -- & --
  & $<-0.2^*$; $>0^*$ & $< 0.5^*$ 
  \\

Idling  
  &-- & --
  & $< 25$ &-- & --
  & --&--
  \\

Start-up/Shut-down 
  & -- & --
  & -- & -- & $<25$
  & -- & --
  \\

High pitch  
  & -- & --
  & -- & -- & --
  & $>30$ & --
  \\

\hline
\end{tabular}
}
\caption{Baseline filters applied to WT A07. Mean, Std, and Min refer to the 10-minute-averaged statistics used for each variable. Filters marked with * are WT-specific.}
\label{tab:Table_2}
\end{table}

\begin{figure}[htbp]
    \centering
    \includegraphics[width=1\textwidth]{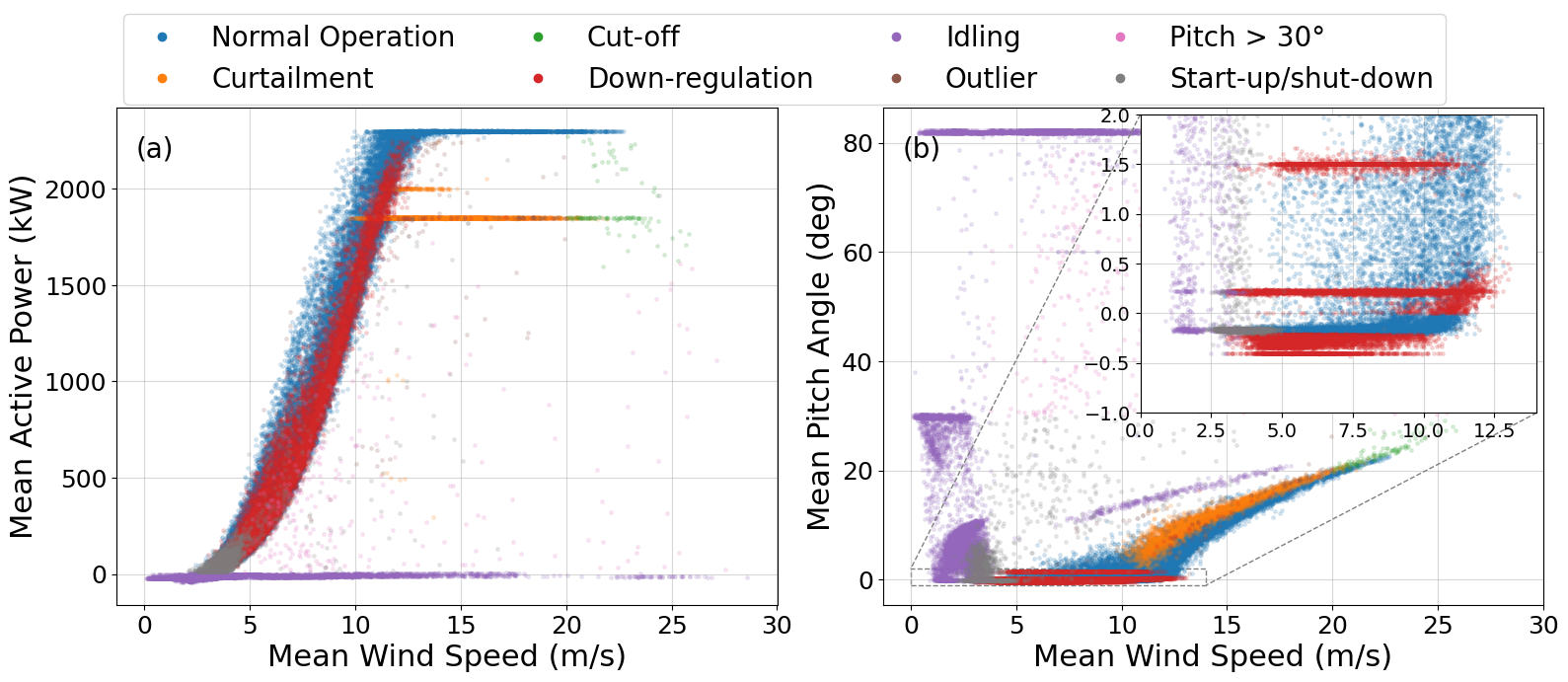}
    \caption{Baseline filtering of WT A07. The filtered and the normal operation data (in blue) are shown in the power curve (a) and in the pitch curve (b).}
    \label{fig:baseline_filtering}
\end{figure}

As an example, the output of baseline filtering of WT A07 is shown in Figure \ref{fig:baseline_filtering}. The data points highlighted in red correspond to down-regulation, toggled during the aforementioned axial-induction control field tests. Not only is this data challenging to detect by visual inspection of the power curve (Figure \ref{fig:baseline_filtering}a), but also by statistical-based methods, as the decrease in power is very limited. On the other hand, by analyzing the blade pitch curve (Figure \ref{fig:baseline_filtering}b), we were able to identify several plateaux of different pitch angles setpoints deviating from the fine pitch of the WT, and define the threshold for the respective filter. While this process was facilitated by the extended field tests' documentation \cite{bossanyi_full-scale_2023}, this is not always the case when analyzing wind farm SCADA data. 

\subsection{Cluster-based filtering}
\label{subsect2.3}

\subsubsection{Clustering algorithms}
\label{sub:clustering_algorithms}

For the cluster-based filtering, the following algorithms were used and compared: i) GMM, ii) DBSCAN, and iii) HDBSCAN. GMM is a probabilistic clustering method that fits multiple Gaussian distributions to the data and estimates the probability that each point belongs to each cluster. By considering both the mean and covariance of the input features, it can capture the underlying correlation between SCADA channels. However, the number of clusters needs to be specified as input. On the other hand, DBSCAN \cite{dbscan} is a density-based clustering algorithm, able to assign data points to clusters of any shape, automatically selecting the optimal number of clusters, according to two input parameters: the minimum number of points (\textit{MinPts}) and the maximum distance ($\epsilon$) between two neighboring points. Both parameters control the detection of noise, i.e., data points not belonging to any cluster, and the expected density of the clusters. Therefore, they can be fine-tuned to detect sparse outliers in SCADA data and to identify dense groups of outliers, such as curtailments. However, as the expected density is fixed, DBSCAN cannot effectively detect clusters of variable density. To overcome this limitation, the use of HDBSCAN \cite{hbscan} is proposed in this study. It is a generalization of DBSCAN to a hierarchical algorithm that can be applied over a wide range of $\epsilon$ values and retains the clusters that persist the longest within this range, enabling the identification of clusters with different densities. Instead of $\epsilon$, the minimum size of a cluster (\textit{{min\_cluster\_size}}) needs to be specified. Finally, a combination of different algorithms is also tested, as further explained in Section \ref{sub:2.3.4}.

\subsubsection{Data pre-processing and feature selection}

Before selecting the input features for each model, the timestamps with non-valid entries were removed. Then, the data channels presented in Table \ref{tab:signals} were standardized to zero mean and unit variance, as different scales and unit-values would skew the calculation of covariance and distance-based metrics of the algorithms.  
Then, the same standardized channels were iteratively provided as inputs to the algorithms, systematically discarding features deemed redundant or unnecessary, based on the resulting clusters and domain-based knowledge. 
At each iteration, the quality of the clusters was assessed through visual inspection, supported by the evaluation metrics defined in Section \ref{sec3}. The features corresponding to the best performing models are listed in Table \ref{tab:Table_3}, for each WT and algorithm used.
In most cases, these include, in addition to the mean values of active power (\textit{pow\_mean}) and local wind speed (\textit{ws\_mean}), data channels such as the mean blade pitch angle (\textit{pitch\_mean}), the mean generator speed in revolutions per minute (\textit{rpm\_mean}), and the pitch standard deviation (\textit{pitch\_std}). However, similar to the manual filters applied for the baseline, the input features for cluster-based filtering remained specific to each WT. 

\begin{table}[htbp]
\centering
\resizebox{\textwidth}{!}{%
\begin{tabular}{llcccccccc}
\hline
WT & Model &
ws\_mean & ws\_min &
pow\_mean & pow\_std &
rpm\_mean & rpm\_std &
pitch\_mean & pitch\_std \\
\hline

\multirow{4}{*}{B08}
  & GMM         & X &   &   & X & X &   & X & X \\
  & DBSCAN      & X &   & X & X & X &   & X & X \\
  & HDBSCAN     & X &   & X & X & X &   & X & X \\
  & DBSCAN+GMM  & X &   & X & X & X &   & X & X \\
\hline

\multirow{4}{*}{A07}
  & GMM         &   & X & X &   &   &   & X & X \\
  & DBSCAN      & X &   & X & X & X &   & X & X \\
  & HDBSCAN     & X &   & X &   &   &   & X & X \\
  & DBSCAN+GMM  &   & X & X &   &   &   & X & X \\
\hline

\multirow{4}{*}{B06}
  & GMM         & X &   & X & X & X &   & X & X \\
  & DBSCAN      & X &   & X & X & X &   & X & X \\
  & HDBSCAN     & X &   & X & & X & X & X & X \\
  & DBSCAN+GMM  & X &   & X & X & X &   & X & X  \\
\hline

\end{tabular}
}
\caption{Features of the best performing models for all three WTs.}
\label{tab:Table_3}
\end{table}

\subsubsection{Hyperparameter tuning}

\begin{figure}[htbp]
\centering
\includegraphics[width=0.6\textwidth]{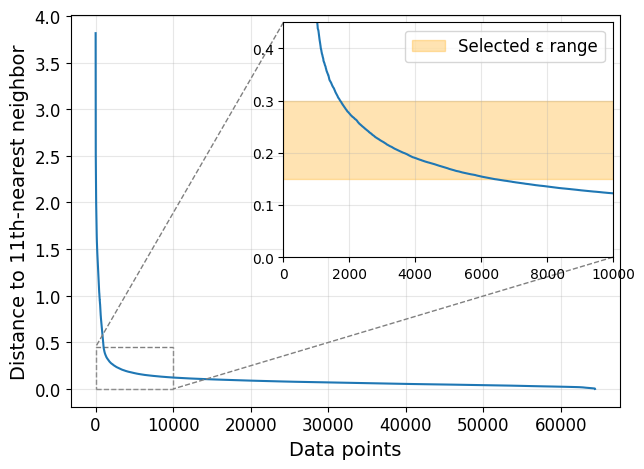}
\caption{k-distance curve for 6-dimension data processed by DBSCAN, for WT A07.}\label{fig:k_distance}
\end{figure}

A similar process was used to tune the hyperparameters of the clustering models, performing a grid search over specific value ranges. Similar to the feature selection, all hyperparameter combinations were evaluated using the metrics defined in Section \ref{sec3}, along with a more detailed visual inspection, if deemed necessary. The selection of hyperparameters is highly dependent on the specific dataset and requires some extent of domain knowledge to estimate the number or size of clusters, according to the model used.
In this study, for the GMM, the number of expected clusters is varied from 5 to 15, based on a preliminary inspection of the operational curves. While we consider this range to cover a realistic number of transients and operational modes for wind farm SCADA filtering, this may not be the case for a different dataset. Therefore, we recommend starting with this range and adjusting it based on the expected types of outliers. 
For DBSCAN, Sander et al. \cite{sander_density-based_1998} suggested to fix \textit{MinPts} to $2 \times dim$, with $dim$ being the dimension of the dataset, i.e., the number of features. 
However, wind farm SCADA data often contain a large number of sparse outliers, so higher values of this parameter may be more effective. In our dataset, we observed that values up to $20 \times dim$ allowed us to capture most of the noise, without excessively increasing the amount of data filtered out.
Therefore, we choose to vary the \textit{min\_samples} between 2 and $20 \times dim$.
To select the value of $\epsilon$, we followed a similar approach to \cite{dbscan} and \cite{sander_density-based_1998}, calculating the k-nearest neighbor distance for each point and analyzing the sorted k-distance curve, with $k=2 \times dim - 1$. Based on the curve, we selected and tested a range of $\epsilon$ values between 0.15 and 0.3, which approximately delimited the elbow of the curve for all WTs, as illustrated in Figure \ref{fig:k_distance} for WT A07.  

Finally, using the same range of \textit{min\_samples}, the \textit{{min\_cluster\_size}} of HDBSCAN is varied from 100 to 500. As this parameter determines the smallest size of grouping considered as a cluster, it needs to be defined according to the quantity of data points processed and the expected cluster size. However, we recommend maintaining this value between 0.1\% and 0.5\% of the number of valid data points, to be able to capture significant periods characterized by specific operational modes.

\subsubsection{Optional pre-filtering}
\label{sub:2.3.4}

With the aim of testing the performance of the models over pre-filtered datasets, the combination of clustering algorithms is also explored. Due to DBSCAN's ability to identify noise, the model is used as a pre-filtering step, before performing the cluster-based filtering. To promote automation and re-usability, the pre-filtering is performed using the same features and hyperparameters for all WTs, fine-tuned with the same methodology described below, but with the main goal of identifying and discarding sparse outliers from the data. Specifically, $\epsilon$ is set to the upper bound of the range defined above, i.e., 0.3, to limit the number of clusters in this preliminary stage, while the recommended value of $2 \times dim = 10$ is used for the \textit{min\_samples}. The mean values of active power, wind speed, pitch angle, and generator speed are used, alongside the power standard deviation. As no significant improvements were observed with DBSCAN and HDBSCAN, which already inherently handle noise, pre-filtering was limited to GMM and is from now on referred to as DBSCAN+GMM.

\subsubsection{Cluster selection}

\begin{figure}[htbp]
    \centering
    \includegraphics[width=0.9\textwidth]{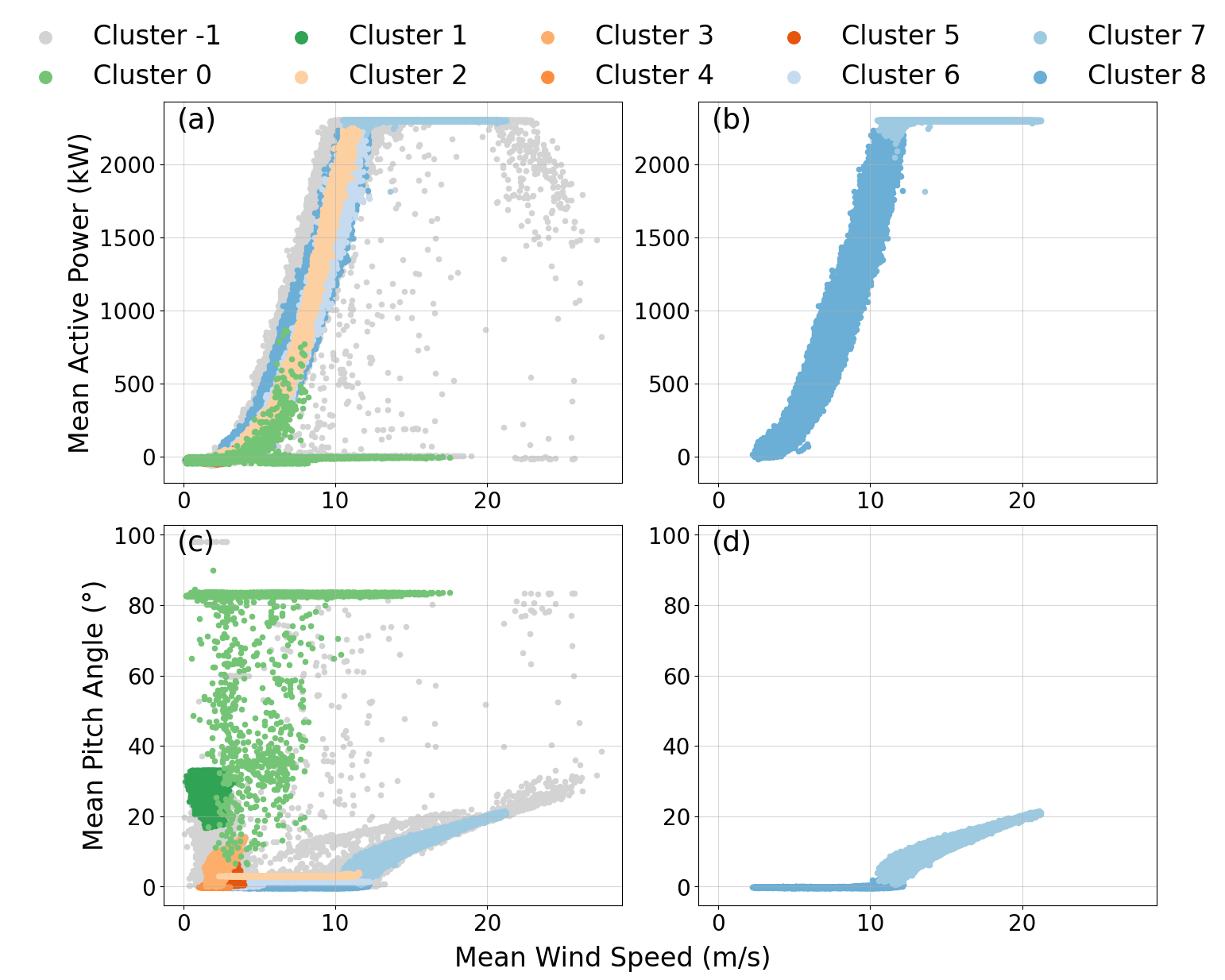}
    \caption{Power and pitch curves of WT B08 with HDBSCAN applied (a, c) and after removing clusters not corresponding to normal operation (b, d).}
    \label{fig:hdbscan_filtering}
\end{figure}

After feature selection and hyperparameter tuning, the clustering itself is performed for each model and WT using the Python open-source library scikit-learn \cite{scikit-learn}. As the models do not automatically distinguish the cluster(s) representing the WT normal operation, a further step is required to identify them through visual inspection of the power and pitch curve, and manually filter out the remaining clusters. An example of cluster selection and filtering is illustrated in Figure \ref{fig:hdbscan_filtering}, where HDBSCAN is applied to WT B08. Clusters 7 and 8 are selected as they represent the below- and above-rated wind speed regimes, respectively, while the remaining clusters, corresponding to noise, idling or start-up (e.g., cluster 0), and down-regulations or operations at non-nominal control setpoints (cluster 2 and 6), are filtered out, using the pitch curve as a visual aid.

\section{Evaluation metrics}
\label{sec3}

As wind farm SCADA data is often unlabeled, it is necessary to define dedicated evaluation metrics to assess and compare the accuracy of different filtering methods. 
As discussed in Section \ref{sec1}, relying on too generalized metrics may limit the interpretability of results, whereas metrics targeted to the wind energy domain should not focus only on specific filtering use cases, such as power curve modeling, but aim to encompass diverse applications. 

For this purpose, we propose evaluation metrics targeted to three WT operational curves: i) the power curve, ii) the pitch curve, and iii) the power-generator speed curve, also known as power-rpm curve. While the first two are widely used when analyzing WT data, the last curve is selected over the rpm-wind speed curve as it provides a clearer distinction among the control settings at which a WT is operated, as further shown below. To define the metrics, we calculate the 10th-90th percentile ranges for the three operational curves, binning by wind speed for the first two curves and by generator rpm for the third. This specific percentile range was generalizable to all curves and WTs in our dataset, ensuring to capture visible outliers in all cases. For the first two curves, wind speed bins of 0.5 m/s are used, while a 50 rpm resolution is selected for the power-rpm curve. Such resolutions enable the capture of transitions of control regimes, e.g., between below- and above-rated wind speeds, while preserving the smoothness of the curves. However, the optimal resolution depends on the dataset used. We recommend using larger bins for data covering shorter time intervals, and adjusting the rpm bins to the rated rpm of the specific generator.

\begin{figure}[htbp]
    \centering
    \centering
    \includegraphics[width=1\textwidth]{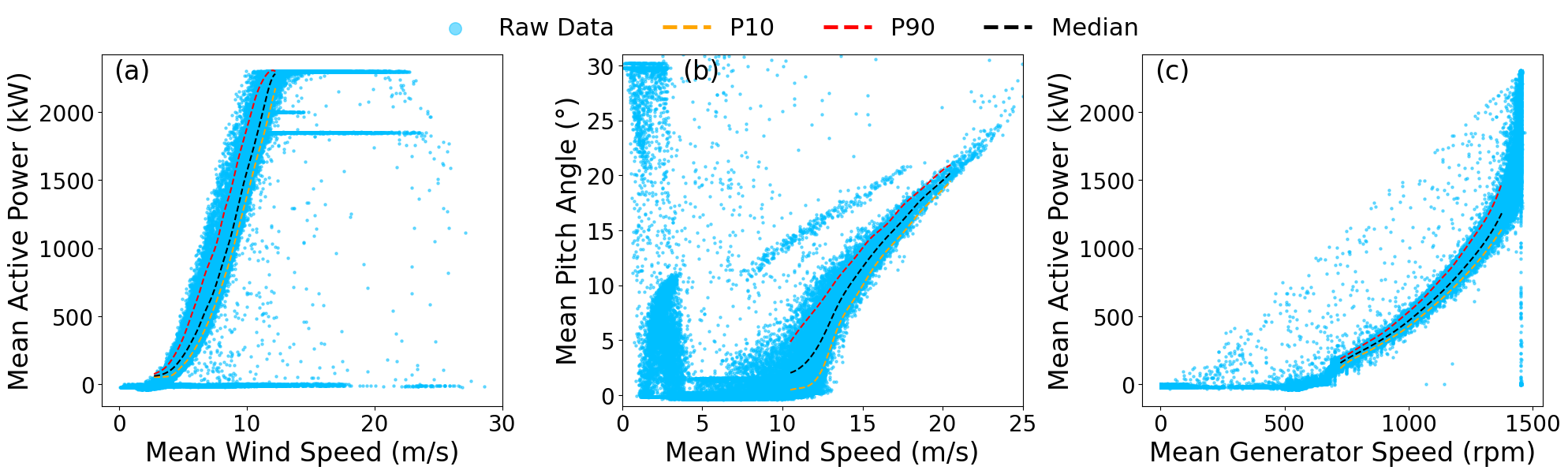}
    \caption{Percentile curves fitted to the power curve (a), pitch curve (b) and power-rpm (c) curve of WT A07.}
    \label{fig:metrcs_curves}
\end{figure}

The binning and curve-fitting are performed independently for each operational curve, following a preliminary filtering of the data, in which we manually filter out the most evident outliers, such as negative min power and curtailments for the power curve, and mean pitch values above 30° for the pitch curve, as they would cause a significant bias for specific bins. Such rough filters are WT-agnostic and are applied merely to define the evaluation metrics. Afterward, we fit the binned percentile curves using cubic spline interpolation, focusing on bins with sufficient measurements (below 21 m/s) and on specific control regimes for each curve: between cut-in and rated wind speeds for the power and power-rpm curve, and above-rated wind speeds for the pitch curve, as illustrated in Figure \ref{fig:metrcs_curves}. 

\begin{figure}[htbp]
\centering
\includegraphics[width=0.7\textwidth]{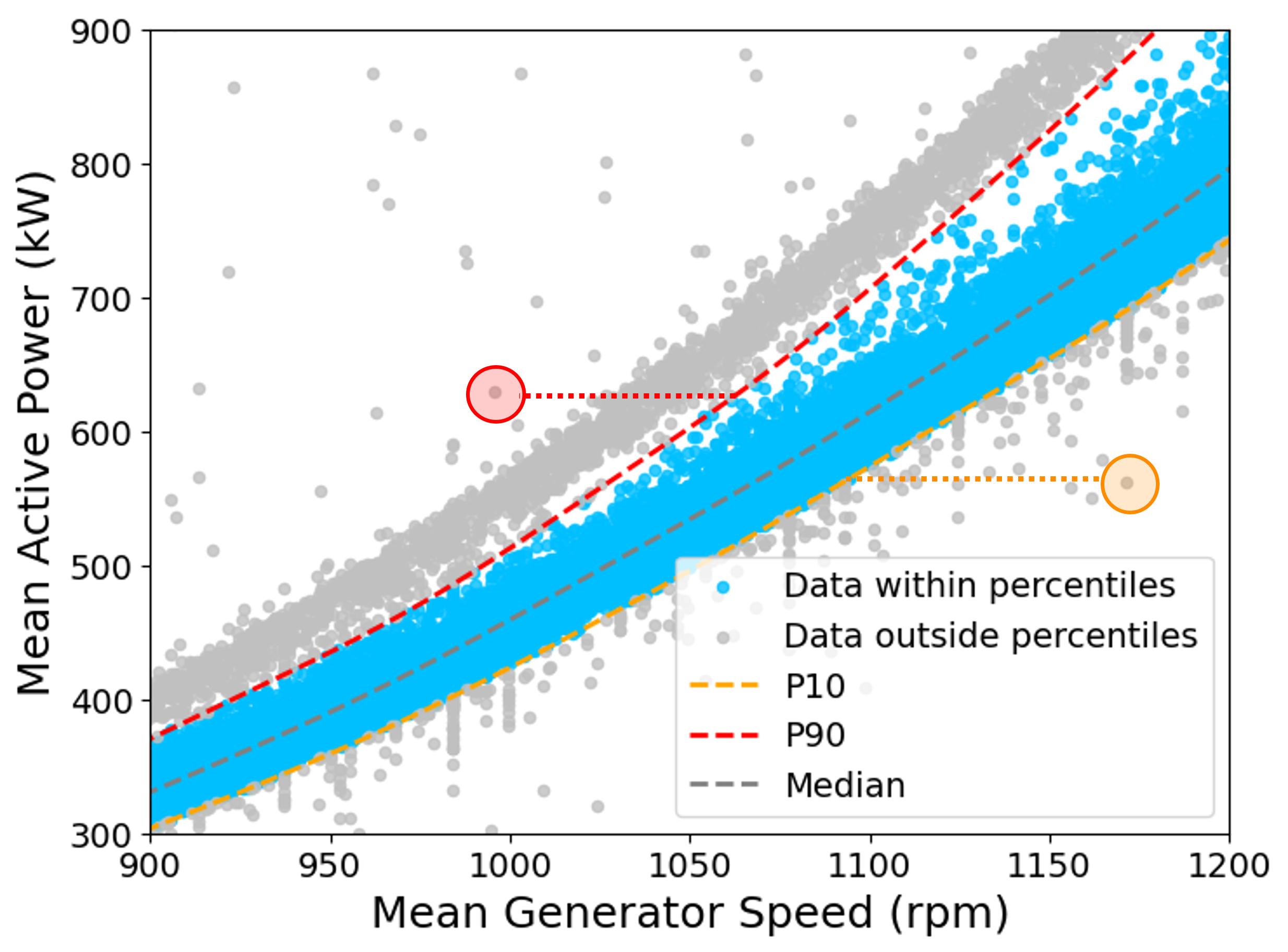}
\caption{Example of evaluation metrics applied to the raw SCADA data of WT B08, analyzing a section of the power-rpm curve. The highlighted points illustrate the computation of distances from percentile ranges. }\label{fig:explain_metrics}
\end{figure}

Such percentile curves are not used as filters themselves, but as statistical indicators for the quality of the filtering, by defining the following metrics:
\begin{itemize}
\item[-] Elimination rate of data points beyond the percentile limits;
\item[-] Decrease in average distance \textit{d} from percentiles;
\item[-] Total retention rate.
\end{itemize}
The first two metrics are applied to all operational curves, while the total retention rate is evaluated over the entire dataset. The elimination rate outside percentiles is computed as $n_{out, filtered}/n_{out, raw}$, where $n_{out, raw}$ denotes the number of points outside the percentiles before the filtering, and $n_{out, filtered}$ is the number of those points that were then filtered out. 
This metric helps penalize the non-detection of stacked outliers, e.g., the down-regulation corresponding to the upper curve outside the percentiles in Figure \ref{fig:explain_metrics}, which is clearly visible in the power–rpm curve. These outliers might be close to the percentile range and therefore not significantly impact the distance-based metric. However, failing to identify them would result in a substantially lower elimination rate outside the percentile range than with accurate filtering, indicating that likely fewer data points deviating from normal operation were filtered out.

The decrease in average distance from the percentile range is calculated as $(d_{raw}-d_{final}) / d_{raw}$, with $d_{raw}$ and $d_{final}$ being the average distance (from the percentile range) of the points outside the percentiles before and after the filtering, respectively. It is mainly used to penalize clear, sparse outliers that are not identified. For example, if several data points such as the ones highlighted in orange and red in Figure \ref{fig:explain_metrics} are not filtered out, this will result in a lower score for this metric.

In addition to these two evaluation metrics applied to the three operational curves, the total retention rate, i.e., the ratio between points that are kept after filtering and the total number of points, is calculated. This or similar metrics are often used in literature \cite{kijanowski_cluster-based_2025, zhao, MORRISON2022473} to quantify the percentage of data, whether or not representing normal operation, removed during the filtering process. A filtering method with a lower retention rate than other methods may indicate that a large share of normal operation data is being unnecessarily discarded, which could be an issue for applications that require a large amount of data.

Overall, the metrics proposed allow estimating the accuracy of wind farm SCADA data filtering, analyzing not only the power curve but also the main variables and curves characterizing WT operation. Using WT-independent pre-filtering and indicators for various outlier types, the metrics can be generalized across different datasets, wind farms, and applications, as further discussed in Section \ref{sec4}. However, the metrics are relative indicators, and thus shall not be used to quantify the accuracy of a given filtering model as an absolute value, but to compare it with a reference baseline or other methods.

\section{Results and discussion}
\label{sec4}

In this section, the results of the cluster-based filtering are presented and compared with the baseline for all three WTs analyzed, using the evaluation metrics presented in Section \ref{sec3}. The strengths and limitations of the proposed methods are then discussed, along with their potential applications. 

\subsection{Filtering results}

\begin{figure}[htbp]
\centering
\includegraphics[width=1\textwidth]{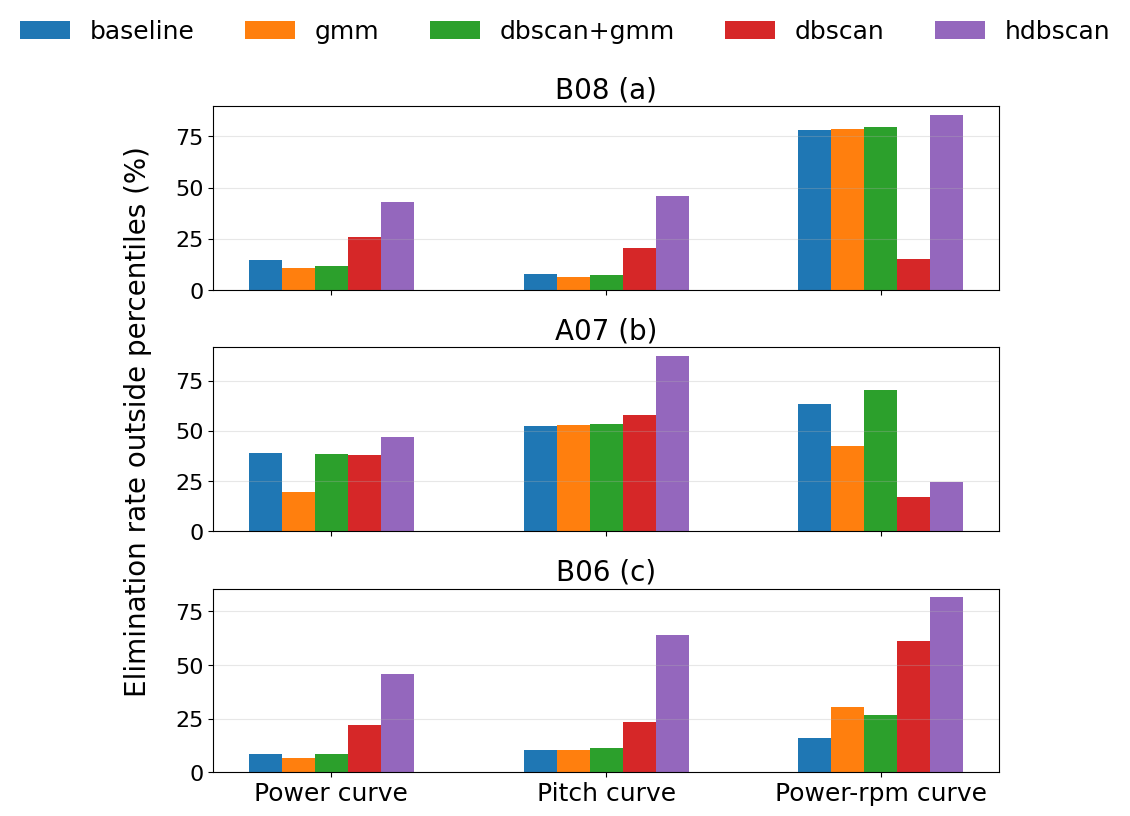}
\caption{Elimination rates for the points outside the percentile range for the power, the pitch, and the power-generator speed (-rpm) curve of turbine B08 (a), A07 (b) and B06 (c), for the baseline and the clustering methods proposed.}\label{fig:points}
\end{figure}

As shown in Figure \ref{fig:points}, elimination rates outside the percentiles vary considerably across WTs, as datasets differ in size and contain different numbers of sparse and stacked outliers, which mainly depend on how the WTs were operated. For example, A07 presents higher elimination rates in the power and pitch curve compared to B08 and B06, as the data covers a shorter period of time, characterized by both curtailments and down-regulations for extended periods. This high proportion of data deviating from normal operation makes A07 the most challenging dataset to filter. In fact, we observe that GMM, both with and without DBSCAN-based pre-filtering (DBSCAN+GMM), generally achieves metrics scores similar to the baseline for WT B08 and B06. However, GMM has considerably worse elimination rates than the baseline for WT A07, unless pre-filtering is performed. 

\begin{figure}[htbp]
    \centering
    \subfigure{
        \includegraphics[width=0.47\textwidth]{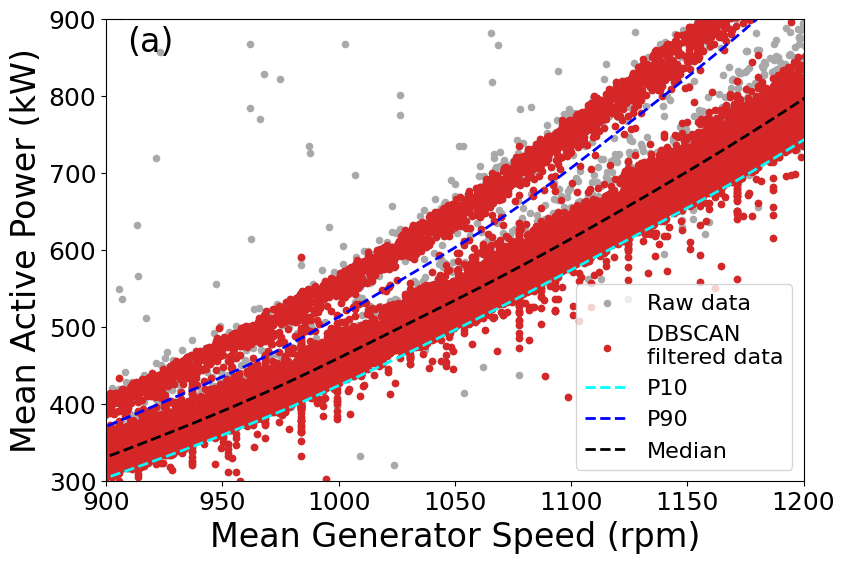}
    }
    \hfill
    \subfigure{
        \includegraphics[width=0.47\textwidth]{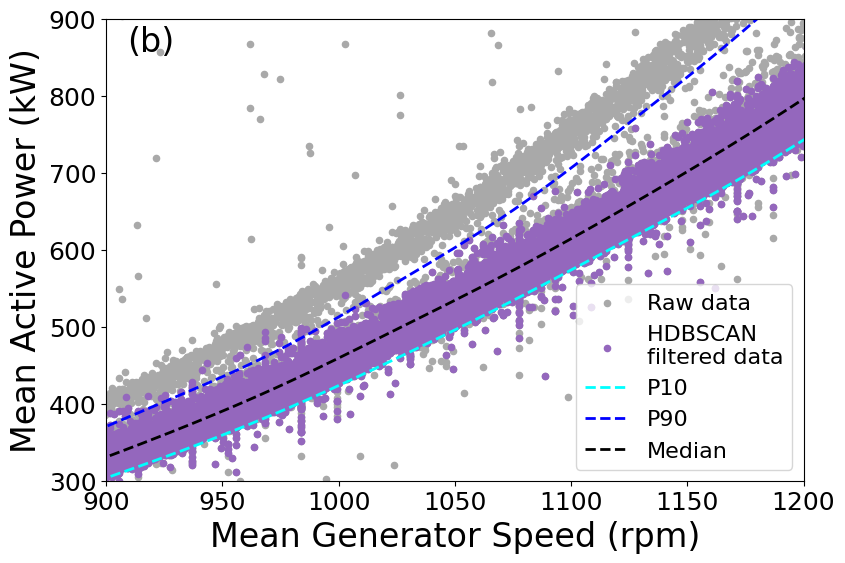}
    }
    \caption{DBSCAN (a) and HDBSCAN-filtered data (b) for WT B08 power-rpm curve, limited between 900 and 1200 rpm. The raw data is displayed in gray for both figures.}
    \label{fig:b08_downregulation}
\end{figure}

Similarly, DBSCAN and HDBSCAN, which overall perform well on this first metric, fail to detect the down-regulation of A07, leading to low scores on the power-rpm curve. This may be explained by the high density of down-regulation data points for this turbine, which density-based algorithms struggle to differentiate from normal operation. On the other hand, while the down-regulation of WT B08 is not detected either by DBSCAN, it is instead identified by HDBSCAN, which, in this case, leverages its adaptability to variable-density clusters. This is clearly visible in Figure \ref{fig:b08_downregulation}, with a zoomed‑in view of the power-rpm curve to emphasize the two operating regimes,  i.e., down-regulated (upper curve) and base control (lower), that DBSCAN fails to distinguish.

\begin{figure}[htbp]
\centering
\includegraphics[width=1\textwidth]{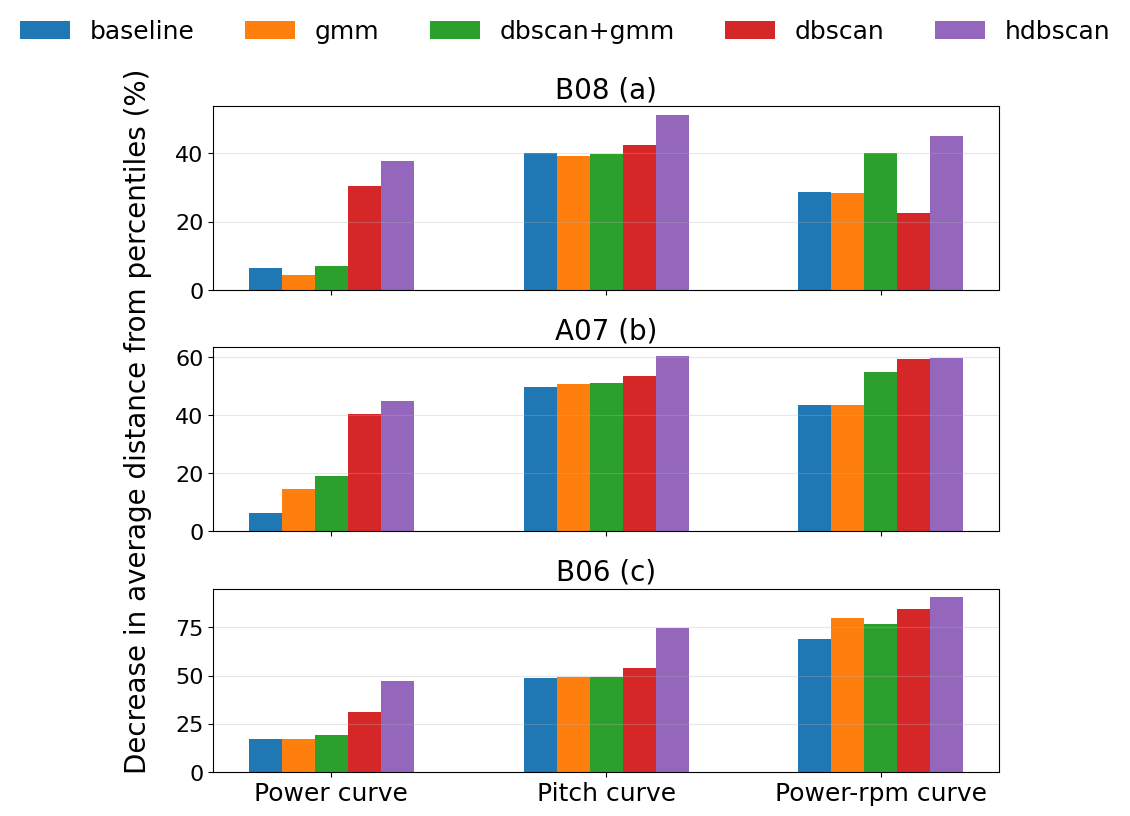}
\caption{Decrease in average distance from the percentile range for the power, the pitch, and the power-generator speed (-rpm) curve of turbine B08 (a), A07 (b), and B06 (c), for the baseline and the clustering methods proposed.}\label{fig:avg_distance}
\end{figure}

With respect to the decrease in average distance from the percentiles, which is mainly used to evaluate the filtering of sparse outliers, the values again vary considerably from WT to WT, due to the differences in the datasets, as shown in Figure \ref{fig:avg_distance}. However, we observe similar trends in the results across operational curves and WTs, with HDBSCAN presenting the largest decrease in average distance, followed by DBSCAN in most cases. This is likely due to the noise-handling mechanism embedded in these algorithms, which facilitates the detection of sparse outliers. However, this may also lead to labeling as noise data points that do not clearly deviate from normal operation but exhibit low density, for example, due to high turbulence intensity at the time of recording. 
This is the case especially for HDBSCAN, which tends to retain mainly the high-density points concentrated around the median and across transitions of control regimes, resulting in more compact operational curves than the other models, particularly for WT B06, as illustrated in Figure \ref{fig:b06_filtered} for the power curves. The pitch and power-rpm curves for the same WT can be found in Appendix \ref{appendix_a}. With regard to GMM, the model achieves values similar to the baseline for this metric, with pre-filtering providing marginal improvements across most WTs and curves.  

\begin{figure}[htbp]
\centering
\includegraphics[width=1\textwidth]{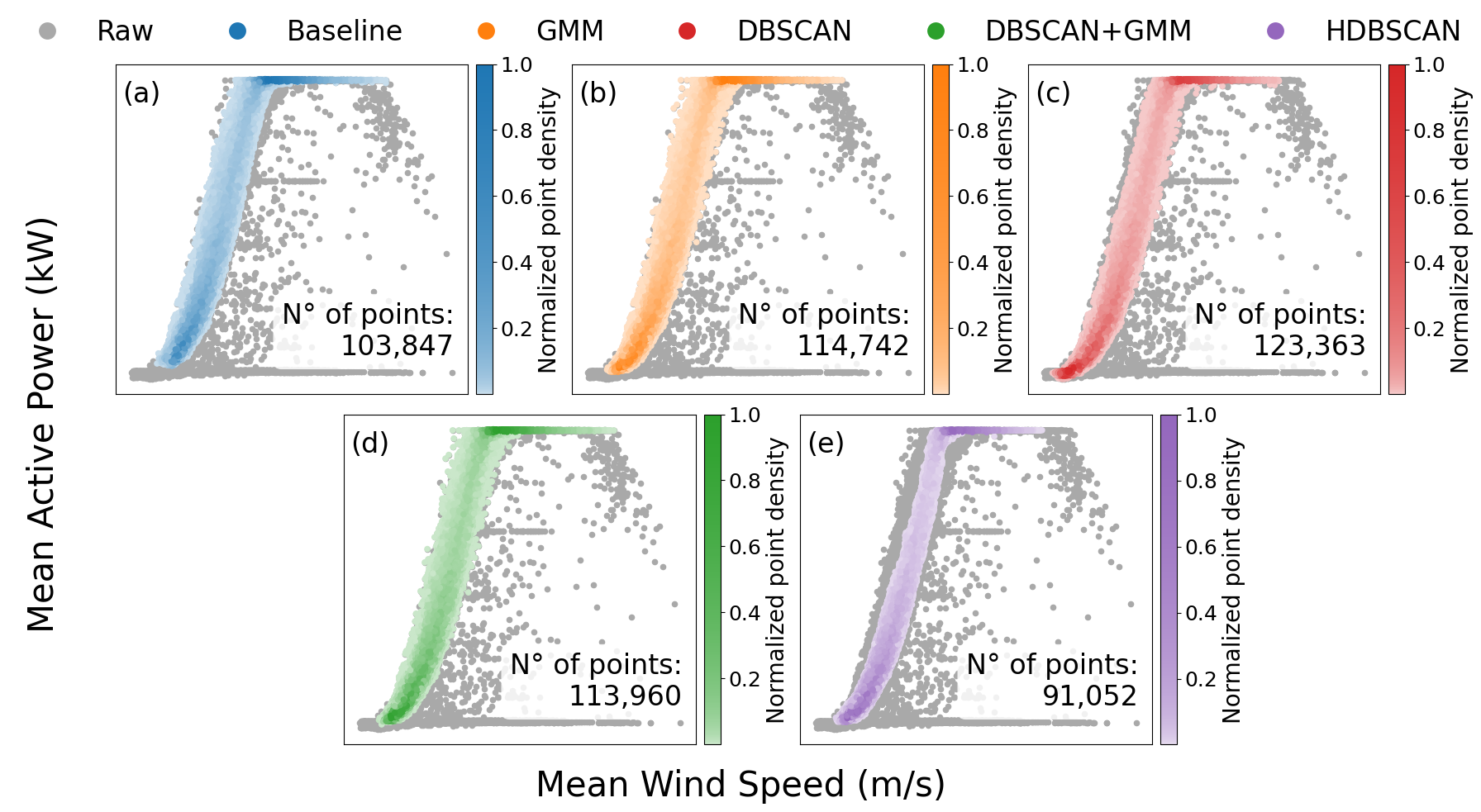}
\caption{Manual- (a), GMM- (b), DBSCAN (c), DBSCAN+GMM- (d) and HDBSCAN-filtered (e) data for WT B06 power curve. Raw data is shown in gray, while filtered data points are colored by point density, computed using two-dimensional binning. }\label{fig:b06_filtered}
\end{figure}

\begin{table}[htbp]
\centering
\begin{tabular}{l ccc}
\hline
\textbf{Model} & \multicolumn{3}{c}{\textbf{Retention Rate (\%)}} \\
               & \textbf{B08} & \textbf{A07} & \textbf{B06} \\
\hline
Baseline   & 68.6 & 49.9 & 73.1 \\
GMM        & 74.1 & 68.4 & 80.7 \\
DBSCAN+GMM & 74.1 & 54.1 & 80.2 \\
DBSCAN     & 76.3 & 70.2 & 75.2 \\
HDBSCAN    & 60.0 & 61.8 & 64.1 \\
\hline
\end{tabular}
\caption{Total retention rates of turbines B08, A07 and B06, for the different filtering models.}
\label{tab:elimination}
\end{table}

\newpage

While the first two metrics provide an estimation of filtering accuracy, it is important to additionally quantify the amount of data retained from the original datasets. The total retention rates of the cluster-based filtering methods are generally higher than those of the baseline filtering, except for HDBSCAN, which presents considerably lower scores for WT B08 and B06, as shown in Table \ref{tab:elimination}. 
However, in most cases, cluster-based methods achieve similar or better accuracy than the baseline, indicating that retention rate can be improved without compromising accuracy. This suggests that rule-based filters often remove more data than necessary, as they impose strict inclusion/exclusion thresholds defined by domain knowledge and visual inspection of the data, whereas clustering algorithms analyze and capture feature correlations and statistical distributions for each dataset. However, all evaluation metrics should be analyzed jointly to ensure a meaningful interpretation. For example, the baseline and the DBSCAN+GMM filtering present the lower total retention rates for WT A07, but mainly because they identified and filtered out down-regulations, as evidenced by the high elimination rate outside the percentiles in the power-rpm curve, unlike the other models.  

\subsection{Model comparison}

In general, all models were able to identify the most evident outliers, such as idling periods, curtailments, and start-up / shut-down transients. GMM performed similarly to the baseline on the first two metrics but consistently presented higher retention rates across the three WTs. The algorithm can capture both stacked outliers, by analyzing the correlations between SCADA channels that lead to the corresponding anomalies or operational modes, and sparse outliers, identified as points with a low probability of belonging to any cluster. This is in line with previous studies, where GMM-based methods achieved good accuracy while preserving the statistical characteristics of wind speeds \cite{MORRISON2022473}.
Applying DBSCAN pre-filtering before GMM clustering leads to only marginal improvements in filtering accuracy, which are not consistent across all turbines and operational curves. This is likely because GMM can effectively identify noise even when applied to raw data. 

On the other hand, DBSCAN proved even more effective at removing sparse outliers, achieving a higher decrease in average distances from the percentiles than GMM and the baseline. 
Still, it struggles to detect non-evident stacked outliers, such as down-regulations or non-nominal control setpoints, because it operates with a fixed expected density. These outliers are found only when a high density is set, but this results in excessive data segmentation and an excessive number of clusters. To overcome this issue, using HDBSCAN is recommended. It works similarly to DBSCAN but, thanks to its hierarchical approach for variable cluster densities, was generally able to identify all stacked outliers, without compromising the distinction of the main clusters corresponding to normal operation. Besides, among all methods, HDBSCAN achieves the highest accuracy in the first two metrics in most cases. Yet, it also tends to filter a large portion of the data, retaining only points that are concentrated around the median operational curves.

\subsection{Potential applications}

Overall, the results show that no model consistently outperforms all others for SCADA data filtering. Instead, each algorithm has its own strengths and limitations, and performs differently depending on the potential filtering application; therefore, the model choice should depend on the specific end use. 
 Specifically, there is an evident trade-off between the total retention rate and how strict the filtering approach is. Stricter filtering leads to lower retention rates, resulting in smaller datasets, but more concentrated around the median operational curves. 
In practice, due to the high retention rates, GMM could be used for applications where the volume of retained data is important. This includes, for instance, training ML models (e.g., power forecasting, structural loads forecasting), where large datasets are typically required; or wake-effect studies and directional power curve modeling, where it is important to preserve sufficient data points for each wind direction sector. On the other hand, the use of HDBSCAN could be targeted at applications where high accuracy is required and in which dataset size is not a priority, such as modeling the power curve of a WT or, more generally, normal behavior modeling across all operational curves. 

Consequently, choosing the appropriate filtering model requires combining multiple evaluation metrics, such as those proposed in this work, to assess both the detection of different outlier types, as well as the capability to retain data, enabling a comprehensive comparison on unlabeled data.
Moreover, the results highlight that metrics focusing exclusively on the power curve may be misleading. In fact, in several instances, e.g., DBSCAN applied to WT B08, a model achieved high accuracy on the power and pitch curves, but not on the power-rpm curve, as downregulation, which may not be clearly visible in the power curve, was not detected. This emphasizes once more the need for a multivariate approach when dealing with wind farm SCADA data, as mean power and wind speed are insufficient to fully characterize WT operation. In this study, we observed that using SCADA channels, such as the mean pitch angle and generator speed, and the pitch and power standard deviation, can improve filtering accuracy, but additional channels could be relevant, depending on the dataset. This aligns with the findings in \cite{astolfi}, which highlight the benefits of considering blade pitch and generator speed in SCADA data analysis, as well as including additional statistics beyond average values. 

Finally, while in this study we focused on identifying normal operation, the proposed methodology could be extended to a classification problem, aiming to identify all remaining clusters, corresponding to different types of anomalies and operational regimes. This could be used, for example, to isolate start-up / shut-down transients, operations at different control setpoints, or to analyze the cause of possible deviations from the expected performance. Future work could extend the methodology by combining it with supervised learning models for cluster recognition. 

\subsection{Limitations}

Among the limitations of the study is the extent to which these findings can be generalized to other datasets and wind farms. In fact, several considerations and parameters worked well for our datasets; however, since SCADA data are very specific to how WTs are operated, they may be less effective in other cases. Nevertheless, the study's applicability is evidenced by applying the methodology to three WTs with different characteristics and by providing recommendations and guidelines for extrapolating the approach to other datasets.

An additional limitation concerns the practicality and efficiency of implementing the cluster-based approaches and replicating them across other wind turbines. Cluster-based methods don't require manually deriving thresholds from a detailed data analysis, which is instead performed to define rule-based filters for each WT in the manual filtering approach. However, feature selection and hyperparameter tuning can still be cumbersome, as multiple combinations must be tested and evaluated. The main bottleneck of the process is the cluster selection step, which requires visual inspection of the power and pitch curves to retain only the clusters corresponding to normal operation, before evaluating filtering accuracy using the metrics. Moreover, we found that, in our case, the feature sets and hyperparameters that worked well for one WT were often not directly transferable to others. As a result, the feature‑selection and tuning procedure had to be re‑performed for each turbine. To address these issues, the manual cluster-selection step could be replaced with approximate filters or image-processing techniques to automatically identify normal-operation clusters from the algorithms' output clusters. In this way, rather than visually inspecting the data for each feature and hyperparameter combination to isolate the normal operation cluster, the metrics' scores could be computed directly. This would allow us to perform a fully automated grid search over the different combinations to quickly retrieve the optimal configuration, thereby increasing the automation of the model calibration pipeline.

Finally, a potential challenge with fully replacing manual filtering with cluster-based filtering is the reduced visibility and explainability of why specific data points are filtered out. This can be mitigated by using the cluster-based approach as a decision-support tool rather than a decision-making tool, keeping an expert in the loop to verify and acknowledge the identified clusters. 

\section{Conclusion}
\label{sec5}

Multiple clustering algorithms were applied in this paper to filter SCADA data from three offshore WTs, benchmarking their results against manual filtering by visual inspection. Additionally, to evaluate and compare filtering accuracy on unlabeled data, we proposed evaluation metrics targeted to WT operational curves and multiple types of outliers, suitable for different applications. 

The results show that in most cases, cluster-based methods achieve higher filtering accuracy than manual filtering, including in the presence of subtle and challenging outliers. However, model calibration and generalization across WTs are not trivial and require domain-based knowledge, as the optimal input features and hyperparameters depend on dataset characteristics. Besides, the comparison of the clustering algorithms indicates that no single method consistently outperforms the others; instead, the choice of the most suitable model depends on the specific application for which filtering is performed. For instance, due to the high retention rates, GMM is well-suited to applications where preserving as much data as possible is a priority. In contrast, HDBSCAN is recommended for stricter filtering, and when the size of the filtered dataset is of secondary importance. These results support the use of the proposed evaluation metrics, which facilitate the comparison of different filtering approaches on unlabeled data, accounting for both outlier detection capability (for sparse and stacked outliers) and data retention, while considering the multivariate nature of SCADA data.

\section*{CRediT authorship contribution statement}

\textbf{Nicolò Italiano:} Writing - Original Draft, Conceptualization, Methodology, Software, Visualization. \textbf{Vasilis Pettas:} Writing - Review \& Editing, Conceptualization, Methodology. \textbf{Tuhfe Göçmen:} Writing - Review \& Editing, Conceptualization, Methodology, Supervision. \textbf{Nicolaos A. Cutululis:} Writing - Review \& Editing, Supervision, Project administration.

\section*{Funding sources}

This research has been supported by the Danmarks Frie Forskningsfond, DFF (grant no. 1127-00188B), under the project “Integrated Design of Offshore Wind Power Plants”, and by the
TWAIN project, which has received funding from the European Union’s Horizon Europe research
and innovation program under grant agreement No. 101122194.

\section*{Acknowledgments}

Special thanks are due to Anders Sommer from Vattenfall for providing the SCADA data used in this study, and to Ebba Dellwik and Leonardo Alcayaga for thier careful review of the manuscript and constructive feedback.

\newpage
\appendix
\section{Appendix: Filtered pitch and power-rpm curves}
\label{appendix_a}
\setcounter{figure}{0}
\begin{figure}[htbp]
\centering
\includegraphics[width=0.85\textwidth]{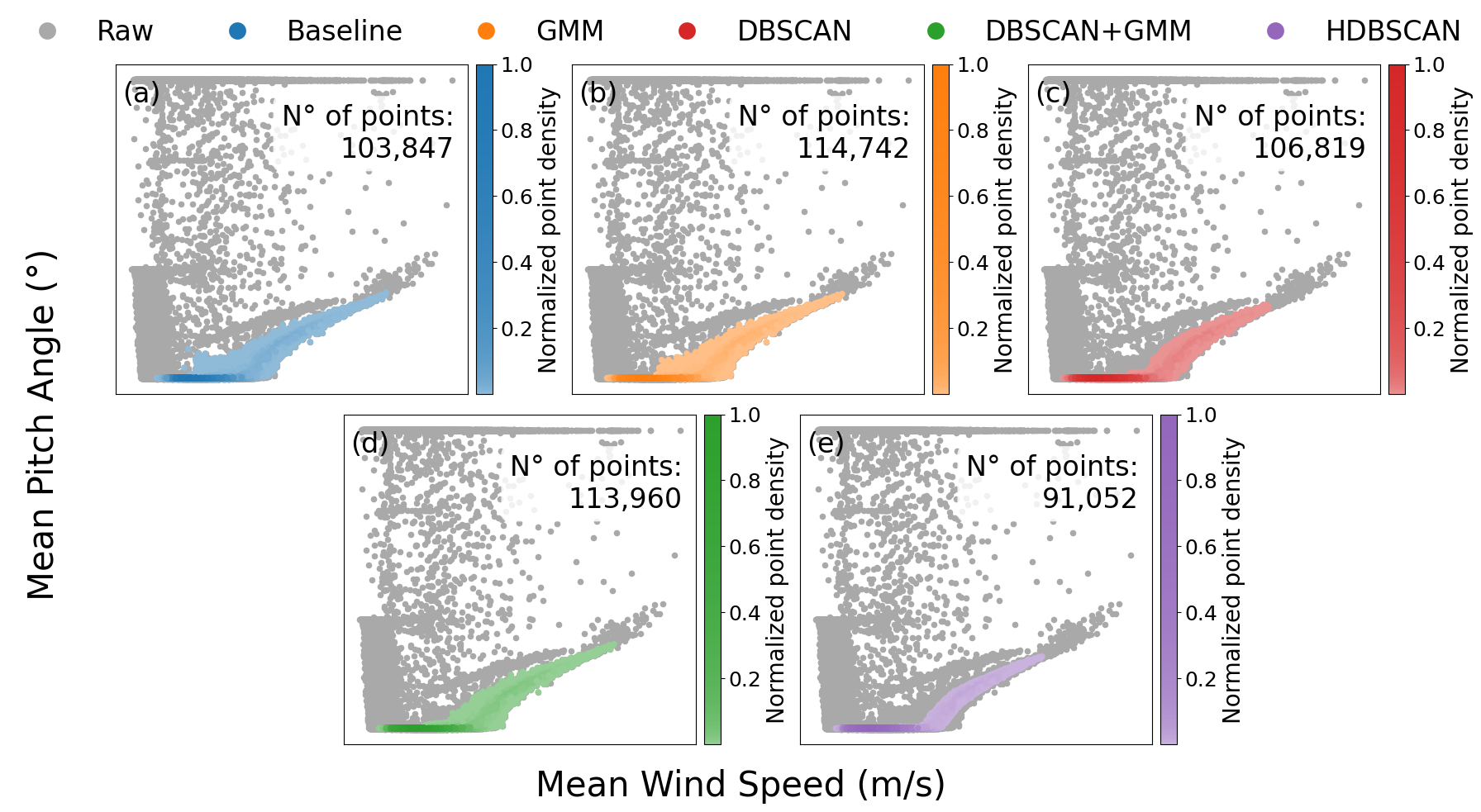}
\caption{Manual- (a), GMM- (b), DBSCAN (c), DBSCAN+GMM- (d) and HDBSCAN-filtered (e) data for WT B06 pitch curve. Raw data is shown in gray, while filtered data points are colored by point density, computed using two-dimensional binning. }\label{fig:five_pitch}
\end{figure}

\begin{figure}[h!]
\centering
\includegraphics[width=0.85\textwidth]{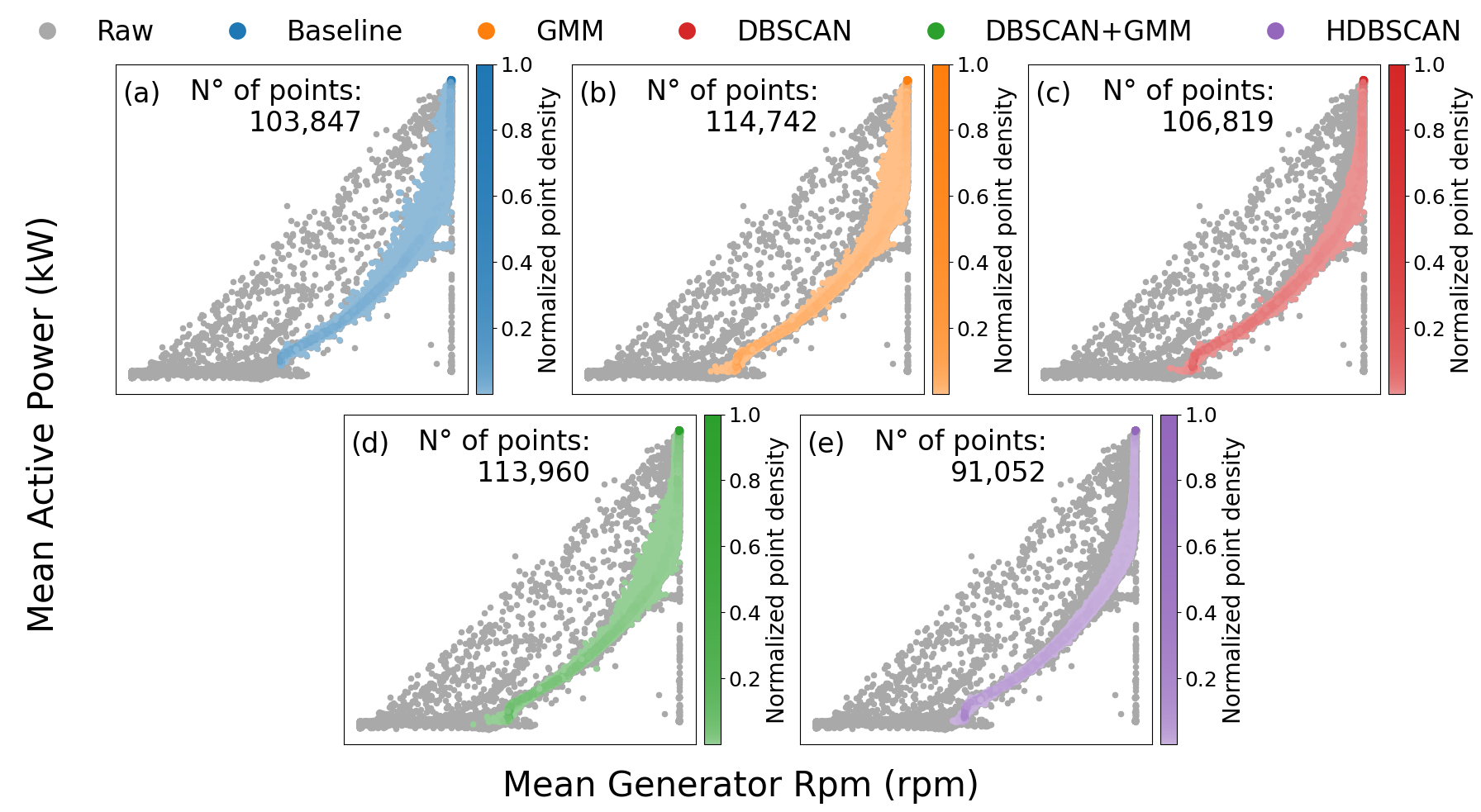}
\caption{Manual- (a), GMM- (b), DBSCAN (c), DBSCAN+GMM- (d) and HDBSCAN-filtered (e) data for WT B06 power-generator speed (-rpm) curve. Raw data is shown in gray, while filtered data points are colored by point density, computed using two-dimensional binning. }\label{fig:five_rpm}
\end{figure}

\newpage
\bibliographystyle{unsrtnat} 


\begin{thebibliography}{10}
\expandafter\ifx\csname url\endcsname\relax
  \def\url#1{\texttt{#1}}\fi
\expandafter\ifx\csname urlprefix\endcsname\relax\def\urlprefix{URL }\fi
\expandafter\ifx\csname href\endcsname\relax
  \def\href#1#2{#2} \def\path#1{#1}\fi

\bibitem{pandit_scada_2023}
R.~Pandit, D.~Astolfi, J.~Hong, D.~Infield, M.~Santos, {SCADA} data for wind turbine data-driven condition/performance monitoring: {A} review on state-of-art, challenges and future trends, Wind Engineering 47~(2) (2023) 422--441.
\newblock \href {https://doi.org/10.1177/0309524X221124031} {\path{doi:10.1177/0309524X221124031}}.

\bibitem{wang}
Y.~Wang, M.~Zhang, X.~Ren, X.~Meng, J.~Yu, E.~Wang, J.~Wang, Y.~Ge, Comparison of artificial intelligence-based power curve cleaning algorithms for wind farms, in: 2023 7th International Conference on Power and Energy Engineering (ICPEE), 2023, pp. 329--336.
\newblock \href {https://doi.org/10.1109/ICPEE60001.2023.10453807} {\path{doi:10.1109/ICPEE60001.2023.10453807}}.

\bibitem{kijanowski_cluster-based_2025}
K.~Kijanowski, T.~Barszcz, P.~B. Dao, A {Cluster}-{Based} {Filtering} {Approach} to {SCADA} {Data} {Preprocessing} for {Wind} {Turbine} {Condition} {Monitoring} and {Fault} {Detection}, Energies 18~(22) (2025).
\newblock \href {https://doi.org/10.3390/en18225954} {\path{doi:10.3390/en18225954}}.

\bibitem{zhao}
Y.~Zhao, L.~Ye, W.~Wang, H.~Sun, Y.~Ju, Y.~Tang, Data-driven correction approach to refine power curve of wind farm under wind curtailment, IEEE Transactions on Sustainable Energy 9~(1) (2018) 95--105.
\newblock \href {https://doi.org/10.1109/TSTE.2017.2717021} {\path{doi:10.1109/TSTE.2017.2717021}}.

\bibitem{luo_method_2022}
Z.~Luo, C.~Fang, C.~Liu, S.~Liu, Method for {Cleaning} {Abnormal} {Data} of {Wind} {Turbine} {Power} {Curve} {Based} on {Density} {Clustering} and {Boundary} {Extraction}, IEEE Transactions on Sustainable Energy 13~(2) (2022) 1147--1159, conference Name: IEEE Transactions on Sustainable Energy.
\newblock \href {https://doi.org/10.1109/TSTE.2021.3138757} {\path{doi:10.1109/TSTE.2021.3138757}}.

\bibitem{wang_pre-filtering_2025}
B.~Wang, B.~Zhou, D.~Zhu, M.~Zou, H.~Luo, Pre-{Filtering} {SCADA} {Data} for {Enhanced} {Machine} {Learning}-{Based} {Multivariate} {Power} {Estimation} in {Wind} {Turbines}, Journal of Marine Science and Engineering 13~(3) (2025).
\newblock \href {https://doi.org/10.3390/jmse13030410} {\path{doi:10.3390/jmse13030410}}.

\bibitem{vasquez-rodriguez_anomaly-based_2024}
G.~Vásquez-Rodríguez, J.~Maldonado-Correa, Anomaly-based fault detection in wind turbines using unsupervised learning: a comparative study., IOP Conference Series: Earth and Environmental Science 1370~(1) (2024) 012005.
\newblock \href {https://doi.org/10.1088/1755-1315/1370/1/012005} {\path{doi:10.1088/1755-1315/1370/1/012005}}.

\bibitem{alcayaga_filtering_2020}
L.~Alcayaga, Filtering of pulsed lidar data using spatial information and a clustering algorithm, Atmospheric Measurement Techniques 13~(11) (2020) 6237--6254.
\newblock \href {https://doi.org/10.5194/amt-13-6237-2020} {\path{doi:10.5194/amt-13-6237-2020}}.

\bibitem{long_image-based_2020}
H.~Long, L.~Sang, Z.~Wu, W.~Gu, Image-{Based} {Abnormal} {Data} {Detection} and {Cleaning} {Algorithm} via {Wind} {Power} {Curve}, IEEE Transactions on Sustainable Energy 11~(2) (2020) 938--946.
\newblock \href {https://doi.org/10.1109/TSTE.2019.2914089} {\path{doi:10.1109/TSTE.2019.2914089}}.

\bibitem{wang_fast_2021}
Z.~Wang, L.~Wang, C.~Huang, A {Fast} {Abnormal} {Data} {Cleaning} {Algorithm} for {Performance} {Evaluation} of {Wind} {Turbine}, IEEE Transactions on Instrumentation and Measurement 70 (2021) 1--12.
\newblock \href {https://doi.org/10.1109/TIM.2020.3044719} {\path{doi:10.1109/TIM.2020.3044719}}.

\bibitem{astolfi}
D.~Astolfi, Perspectives on scada data analysis methods for multivariate wind turbine power curve modeling, Machines 9~(5) (2021).
\newblock \href {https://doi.org/10.3390/machines9050100} {\path{doi:10.3390/machines9050100}}.

\bibitem{MORRISON2022473}
R.~Morrison, X.~Liu, Z.~Lin, Anomaly detection in wind turbine scada data for power curve cleaning, Renewable Energy 184 (2022) 473--486.
\newblock \href {https://doi.org/https://doi.org/10.1016/j.renene.2021.11.118} {\path{doi:https://doi.org/10.1016/j.renene.2021.11.118}}.

\bibitem{exploratory}
P.~C. Rodriguez, P.~Marti-Puig, C.~F. Caiafa, M.~Serra-Serra, J.~Cusidó, J.~Solé-Casals, Exploratory analysis of scada data from wind turbines using the k-means clustering algorithm for predictive maintenance purposes, Machines 11~(2) (2023).
\newblock \href {https://doi.org/10.3390/machines11020270} {\path{doi:10.3390/machines11020270}}.

\bibitem{zhang_wind_2024}
S.~Zhang, E.~Robinson, M.~Basu, Wind turbine condition monitoring based on three fitted performance curves, Wind Energy 27~(5) (2024) 429--446.
\newblock \href {https://doi.org/10.1002/we.2859} {\path{doi:10.1002/we.2859}}.

\bibitem{lin_wind_2020}
Z.~Lin, X.~Liu, M.~Collu, Wind power prediction based on high-frequency {SCADA} data along with isolation forest and deep learning neural networks, International Journal of Electrical Power \& Energy Systems 118 (2020) 105835.
\newblock \href {https://doi.org/10.1016/j.ijepes.2020.105835} {\path{doi:10.1016/j.ijepes.2020.105835}}.

\bibitem{he_oct_2024}
G.~He, X.~Gao, L.~Li, P.~Gao, {OCT} monitoring data processing method of laser deep penetration welding based on {HDBSCAN}, Optics \& Laser Technology 179 (2024) 111303.
\newblock \href {https://doi.org/10.1016/j.optlastec.2024.111303} {\path{doi:10.1016/j.optlastec.2024.111303}}.

\bibitem{bossanyi_full-scale_2023}
E.~Bossanyi, G.~C. Larsen, M.~M. Pedersen, Full-scale validation of optimal axial induction control of a row of turbines at {Lillgrund} wind farm, Journal of Physics: Conference Series 2505~(1) (2023) 012042.
\newblock \href {https://doi.org/10.1088/1742-6596/2505/1/012042} {\path{doi:10.1088/1742-6596/2505/1/012042}}.

\bibitem{flasc2026}
NLR, \href{https://github.com/NatLabRockies/flasc}{Flasc. version 2.4.1}, (accessed 5 January 2026) (2026).
\newline\urlprefix\url{https://github.com/NatLabRockies/flasc}

\bibitem{wake_steering}
D.~Siguenza-Alvarado, M.~Harrison, M.~Mohammadi, P.~Vishwakarma, E.~Bossanyi, L.~Landberg, M.~Bastankhah, Assessing engineering wake models against operational data: insights from the {Lillgrund} wind farm wake steering campaign (2026).
\newblock \href {https://doi.org/10.48550/arXiv.2601.21035} {\path{doi:10.48550/arXiv.2601.21035}}.

\bibitem{dbscan}
M.~Ester, H.-P. Kriegel, J.~Sander, X.~Xu, A density-based algorithm for discovering clusters in large spatial databases with noise, in: Proceedings of the {Second} {International} {Conference} on {Knowledge} {Discovery} and {Data} {Mining}, {KDD}'96, AAAI Press, Portland, Oregon, 1996, pp. 226--231.

\bibitem{hbscan}
L.~McInnes, J.~Healy, S.~Astels, hdbscan: Hierarchical density based clustering, Journal of Open Source Software 2~(11) (2017) 205.
\newblock \href {https://doi.org/10.21105/joss.00205} {\path{doi:10.21105/joss.00205}}.

\bibitem{sander_density-based_1998}
J.~Sander, M.~Ester, H.-P. Kriegel, X.~Xu, Density-{Based} {Clustering} in {Spatial} {Databases}: {The} {Algorithm} {GDBSCAN} and {Its} {Applications}, Data Mining and Knowledge Discovery 2~(2) (1998) 169--194.
\newblock \href {https://doi.org/10.1023/A:1009745219419} {\path{doi:10.1023/A:1009745219419}}.

\bibitem{scikit-learn}
F.~Pedregosa, G.~Varoquaux, A.~Gramfort, V.~Michel, B.~Thirion, O.~Grisel, M.~Blondel, P.~Prettenhofer, R.~Weiss, V.~Dubourg, J.~Vanderplas, A.~Passos, D.~Cournapeau, M.~Brucher, M.~Perrot, E.~Duchesnay, Scikit-learn: Machine learning in {P}ython, Journal of Machine Learning Research 12 (2011) 2825--2830.

\end{thebibliography}

\end{document}